\documentclass{article}

\usepackage{PRIMEarxiv}

\usepackage[utf8]{inputenc} % allow utf-8 input
\usepackage[T1]{fontenc}    % use 8-bit T1 fonts
\usepackage{url}            % simple URL typesetting
\usepackage{booktabs}       % professional-quality tables
\usepackage{amsfonts}       % blackboard math symbols
\usepackage{nicefrac}       % compact symbols for 1/2, etc.
\usepackage{microtype}      % microtypography
\usepackage{lipsum}
\usepackage{fancyhdr}       % header
\usepackage{graphicx}       % graphics
\graphicspath{{media/}}     % organize your images and other figures under media/ folder

\title{Surgical Gesture Recognition Based on  Bidirectional Multi-Layer Independently  RNN with Explainable Spatial Feature Extraction
}

\author{
  Dandan Zhang*, Ruoxi Wang*, Benny Lo \\
  
}

\begin{document}
\maketitle

\begin{abstract}

Minimally invasive surgery mainly consists of a series of sub-tasks, which can be decomposed into basic gestures or contexts. As a prerequisite of autonomic operation, surgical gesture recognition can assist motion planning and decision-making, and build up context-aware knowledge to improve the surgical robot control quality. In this work, we aim to develop an effective surgical gesture recognition approach with an explainable feature extraction process.

A   Bidirectional Multi-Layer independently  RNN (BML-indRNN) model is proposed in this paper, while spatial feature extraction is implemented via fine-tuning of a Deep Convolutional Neural Network (DCNN) model constructed based on the VGG architecture. To eliminate the black-box effects of DCNN, Gradient-weighted Class Activation Mapping (Grad-CAM) is employed. It can provide explainable results by showing the regions of the surgical images that have a strong relationship with the surgical gesture classification results.

The proposed method was evaluated based on the suturing task with data obtained from the public available JIGSAWS database. Comparative studies were conducted to verify the proposed framework. Results indicated that the testing accuracy for the suturing task based on our proposed method is 87.13\%, which outperforms most of the state-of-the-art algorithms.
\end{abstract}

\section{Introduction}

In the past few decades, Robot-Assisted Minimally Invasive Surgery (RAMIS) has transformed surgery and brought great benefits to patients, such as reduced recovery time and traumas \cite{bergeles2013passive}. Moreover, it enables surgeons to conduct surgery remotely via a master manipulator\cite{zhang2019handheld,miyazaki2018master,Zhang2020Hamlyn}, which can protect surgeons from radiation and provide ergonomic comfort \cite{zhang2019design,zhang2020ergonomic}.
With RAMIS, the surgical operation data can be recorded from the robotic system and used for surgical gesture segmentation and recognition, surgical skills assessment \cite{lin2006towards,Zhang2020Automatic,zhang2020microsurgical}, robotic assistance \cite{ruurda2002robot,Zhang2018Self,payne2021shared}, and automation \cite{preda2016cognitive,chen2020supervised,yang2017medical}. Since surgical gesture recognition is essentially required for the support of high-level perception of surgical workflow \cite{van2019weakly}, we focus on the implementation of automatic surgical gesture recognition with explainable features in this paper.

\begin{figure}[tb]  
	\centering
	\includegraphics[width = 0.8\hsize]{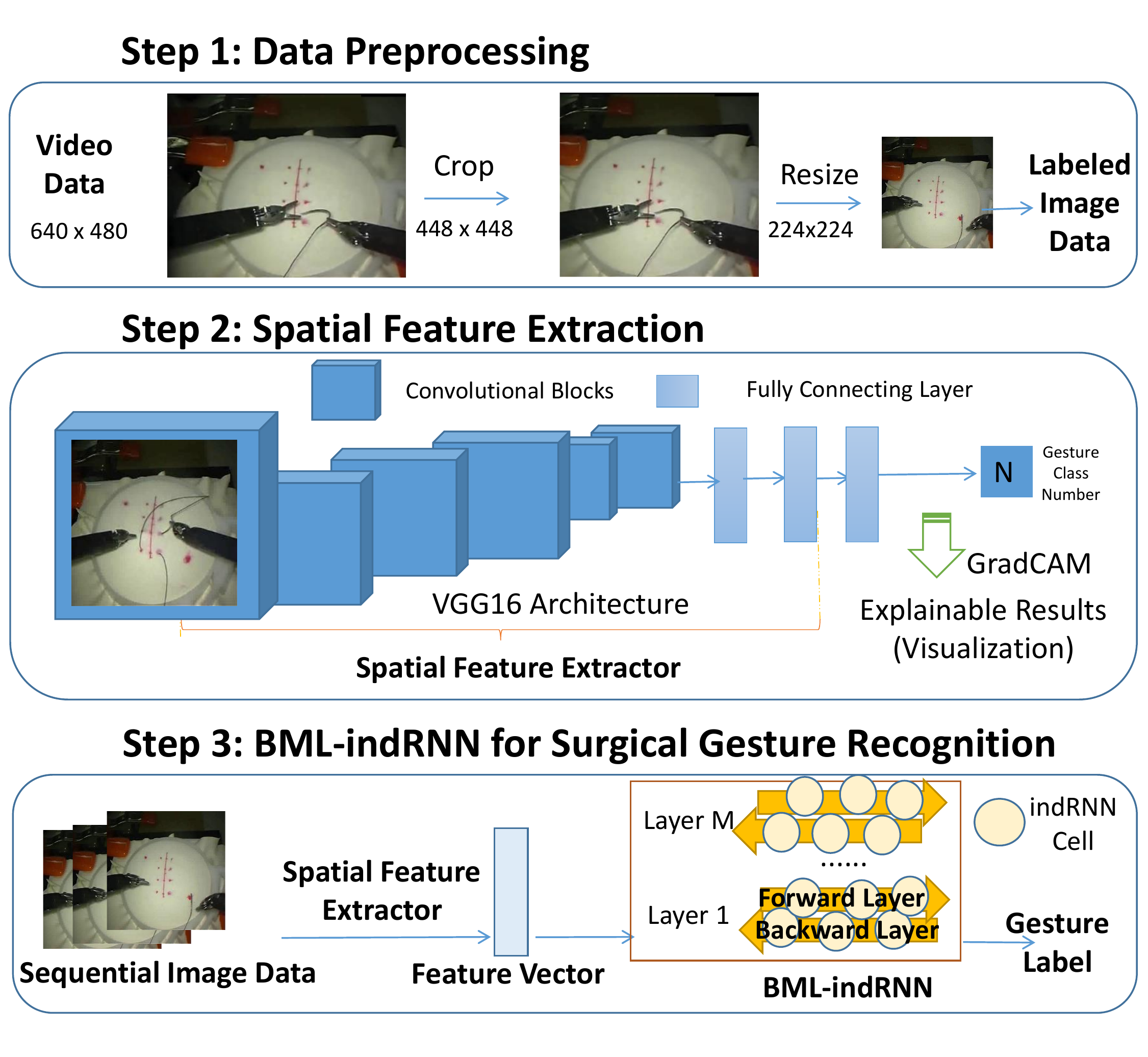}
	\caption{The workflow for the implementation of the proposed method for surgical gesture classification.}
	\label{fig:Overview.pdf}
\end{figure}
%HMM,
To process sequential data for surgical gesture recognition, classical approaches, such as Hidden Markov Model (HMM) \cite{Leong2006HMM} has been widely used. HMM estimates the transition of hidden states to characterize the surgical task sequences. Surgical gestures can be modeled as one or more states of an HMM, while the observations can be modeled differently  \cite{P2-5}. For example, the observations are assumed to be generated from a lower-dimensional latent space using Factor Analyzed HMMs (FA-HMMs) and Switched Linear Dynamical Systems (SLDSs) \cite{varadarajan2011learning}.  Gaussian mixture model (GMM) \cite{van2019weakly,perez2019using} and mixtures of factor analyzers (MFAs) can be used to capture the variability of complex operations as observations. In addition to HMM, Dynamic Time Wrapping \cite{Forestier2011Assessment}, Conditional Random Fields (CRF) \cite{mavroudi2018end}  and other machine learning based time sequences processing technologies have been utilized for surgical pattern recognition as well.

% and prior knowledge

Deep Neural Networks (DNNs) have emerged with promising results in many applications. Compare to most of the traditional statistical models  for sequential data modeling, DNNs can extract significant features from data automatically without manually selecting features  \cite{ravi2017deep,ermes2008detection}. To incorporate temporal information for analyzing time series data, Long Short-Term Memory (LSTM) has been  employed, which can preserve temporal characteristics of the signals \cite{gers2002applying} by fusing information from previous steps and instantaneous inputs. In addition, Temporal convolutional network (TCN) \cite{lea2016temporal} and 3D convolutional network \cite{funke2019using} have been proved to be promising for sequential data processing as well, which have been used to extract representative visual features for surgical gesture recognition. They have shown an important role in applications such as action recognition, semantic labeling, language or video processing. More recently, Fusion-KV \cite{qin2020temporal} was proposed to combine the vision data and kinematics data recorded from the surgical robotic system for gesture recognition, while Fusion-KV with an attention-based LSTM decoder has been utilized to further enhance the recognition performance \cite{despinoy2015unsupervised}.

In addition to supervised learning based approaches, deep reinforcement learning (RL) has been investigated in gesture recognition, where an agent learned its policy of surgical gesture segmentation and classification by interacting with the surgical data \cite{liu2018deep}. However, the network confusion problem has not been addressed for some less frequent surgical gestures \cite{itzkovich2019using}.    A reinforcement learning and tree search based framework has been proved to be effective for surgical gesture recognition \cite{gao2020automatic}, where the visual features are regarded as the environment states while the determination of the classes for the surgical gesture is regarded as actions.  A tree search algorithm is used to combine the outputs of the policy and the value network to obtain a better performance.

%Explainable results for the feature extraction process are visualized to demonstrate the significance of fine-tuning the spatial feature extraction model. 
The aim of this paper is to propose a  Bidirectional Multi-Layer independently  RNN (BML-indRNN) model with spatial feature extraction for surgical gesture recognition. Apart from accurate surgical gesture recognition, the work also provides explainable results by showing how the neural network classifies the surgical gestures. Considering that video data includes semantic information and can be more intuitive for identifying the context compared to kinematics data, we use video data for the implementation of automatic surgical gesture recognition.   

The details of the paper are described as follows. Firstly, the methodology is introduced in Section II. Secondly, the experimental setup and results analysis are presented in Section III. Finally, conclusions are drawn in Section IV.

\section{Methodology}
\label{sec:context_awareness}

\subsection{Overview}

 Video data of surgical operation includes both spatial and temporal features that are useful for surgical gesture recognition. Deep Convolutional Neural Network (DCNN) has demonstrated promising results in image classification with powerful spatial feature extraction functions. However, it is not suitable for identifying the temporal features of sequential data. Therefore, we intend to combine the advantages of DCNN for spatial feature extraction and RNN for temporal feature extraction, through which the method can enhance the gesture classification accuracy.

The overview of the workflow of the proposed method for surgical gesture classification is shown in Fig. \ref{fig:Overview.pdf}. The first step is data pre-processing, which requires transforming video data into a series of sequential images, cropping and reshaping the frames to the desired sizes and labeling each image to the corresponding surgical gesture class. The second step is spatial feature extraction, which is implemented by using the DCNN model. VGG model, as a typical architecture of DCNN  is used for spatial feature extraction. It can compress the 2D image array to a 1D feature vector while preserving the most useful information before feeding into the RNN. The third step is surgical gesture recognition based on the proposed BML-indRNN model, which extracts the temporal features automatically and predicts the surgical gestures. More details of the processing pipeline are described in the following sections.

\subsection{Database Description}

Since the JIGSAWS database is a publicly available database with several surgical tasks, we chose the suturing task as an example to test the performance of the proposed method in this paper. Video data of the JIGSAWS database is used for validation. The definitions of the surgical gesture for suturing task are shown in Table. \ref{tab:Definition}, which are adapted from \cite{ahmidi2017dataset}. Ten types of surgical gestures are defined in total. Thus, surgical gesture recognition can be formulated as a 10-class classification problem. 

%Based on the surgical gesture defined for the suturing task in  \cite{ahmidi2017dataset}

\begin{table}[!htb]
	\centering
	\caption{Definitions for Surgical Gestures Involved in the Suturing Task}
	\label{tab:Definition}
	\begin{tabular}{c|l}
		\hline\hline
		\textbf{Surgical Task} & \textbf{Surgical Gesture Definition} \\\hline
		& G1: Reaching for needle with right hand \\
		& G2:  Positioning needle  \\
		& G3: Pushing needle through tissue \\
		& G4:  Transferring needle from left to right \\		        
		Suturing & G5: Moving to center with needle in grip \\
		&  G6: Pulling suture with left hand \\
		& G7:  Orienting needle \\
		& G8: Using right hand to help tighten suture \\
		& G9:  Loosening more suture \\
		& G10:  Dropping suture at end and moving to end points \\
		\hline	\hline		
	\end{tabular}
\end{table}

\subsection{Explainable Spatial Feature Extraction}

\subsubsection{Model Construction}
%while VGG model is used to extract the features with dimension of $4096{\times}1$
%\times T

%The video data is consists of $T$ number of RGB images, where $T$ represents the total number of frames for one trial of the operation videos. 

The JIGSAWS has been manually annotated with the ground-truth surgical gesture at frame level \cite{gao2014jhu}. Therefore, the corresponding relationship between each frame and the gesture label can be obtained for model training. The size of the original frame of the video data is $640 \times 480$. The videos are recorded at a frame rate of 30$hz$. During the model training process, the frame for the video data is downsampled with factor $r=5$.  Furthermore, each frame is cropped to have the dimension of $448 \times 448$ pixels and is resized to  $224 \times 224$  pixels to accelerate the network training and inferencing process.

VGG16 model is designed for image classification \cite{simonyan2014very}, which has been proven to be able to classify images accurately from the ImageNet dataset \cite{Deng2009ImageNet}. To extract meaningful features for the state representation of the surgical operation image data, we use the VGG16 model with pre-trained weights on ImageNet for feature extraction. However, the pre-trained model may have a bias when used for feature extraction for surgical gesture classification, since the model is trained for object recognition originally. Therefore, we need to fine-tune the model for effective feature extraction using the task-orientated database to obtain a better state representation of the image data.

\begin{figure}[tb]  
	\centering
	\includegraphics[width = 0.8\hsize]{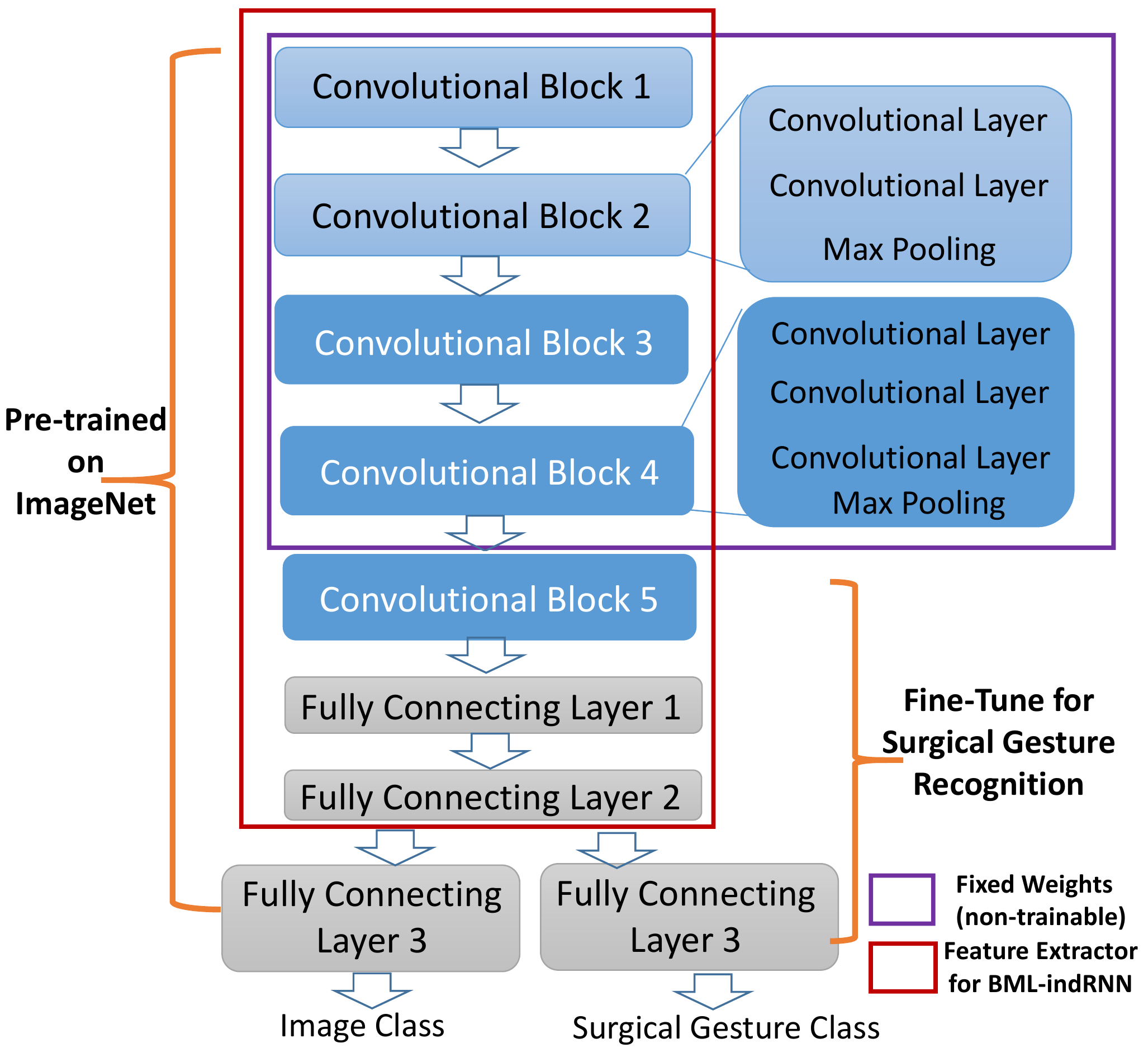}
	\caption{The architecture of the feature extraction model.}
	\label{fig:VGG-Features.pdf}
\end{figure}

The modification of the architecture for spatial feature extraction is shown in Fig. \ref{fig:VGG-Features.pdf}.
The weights of the convolutional blocks and the first fully connecting layer are frozen, which can form a feature extraction model, while the final fully connecting layer of the network for mapping the features to specific classes of the ImageNet is removed. Following that, we add a fully connecting layer with input units of $4096 \times 1$ and output units of $N$ to the end of the model, where $N$ represents the number of surgical gesture classes defined for the specific recognition task.

During model training, the weights of the convolutional block 1 to convolutional block 4 are fixed, while the convolutional block 5 and all the fully connecting layers are trainable for frame-wise surgical gesture classification. Once the new model is obtained, a new feature extraction model can be constructed after removing the final fully connecting layer. The input of the feature extraction model is an image array after pre-processing, while the output of the model is a compact feature vector with the dimension of $4096 \times 1$. The obtained feature vector  can then be fed into the BML-indRNN for surgical gesture classification, as described in the next section.

\subsubsection{Interpretability}

Though DCNN has been proven to have outstanding performance in classification, it has inherent black-box effects. In order to make the model transparent for people to under the reason why the series of frames are classified as a specific gesture, visualization techniques should be used to provide interpretable  results obtained via the model. In this way, the effectiveness of the spatial feature extraction process can be evaluated qualitatively.

Class Activation Map has been developed to show the critical parts that contribute the most to the classification results \cite{meng2019class}\cite{2015Learning}. However, a global activation map layer in the neural network model is necessary, which limits its general applications. Gradient-weighted Class Activation Mapping (Grad-CAM) \cite{selvaraju2017grad} has been proposed, which is able to mitigate the black-box effect of DCNN without re-training the model. The class-specific gradient information can be conveyed to the final convolutional layer of a DCNN, while the critical parts that contribute the most to the classification results can be visualized in the form of  a coarse localization map  with important regions highlighted. Grad-CAM can help users establish trust in predictions from DCNN. Therefore, Grad-CAM is used to provide an explainable spatial feature extraction post-hoc analysis in this paper.

\subsection{BML-indRNN}

\begin{figure*}[tb]  
	\centering
	\includegraphics[width = 1\hsize]{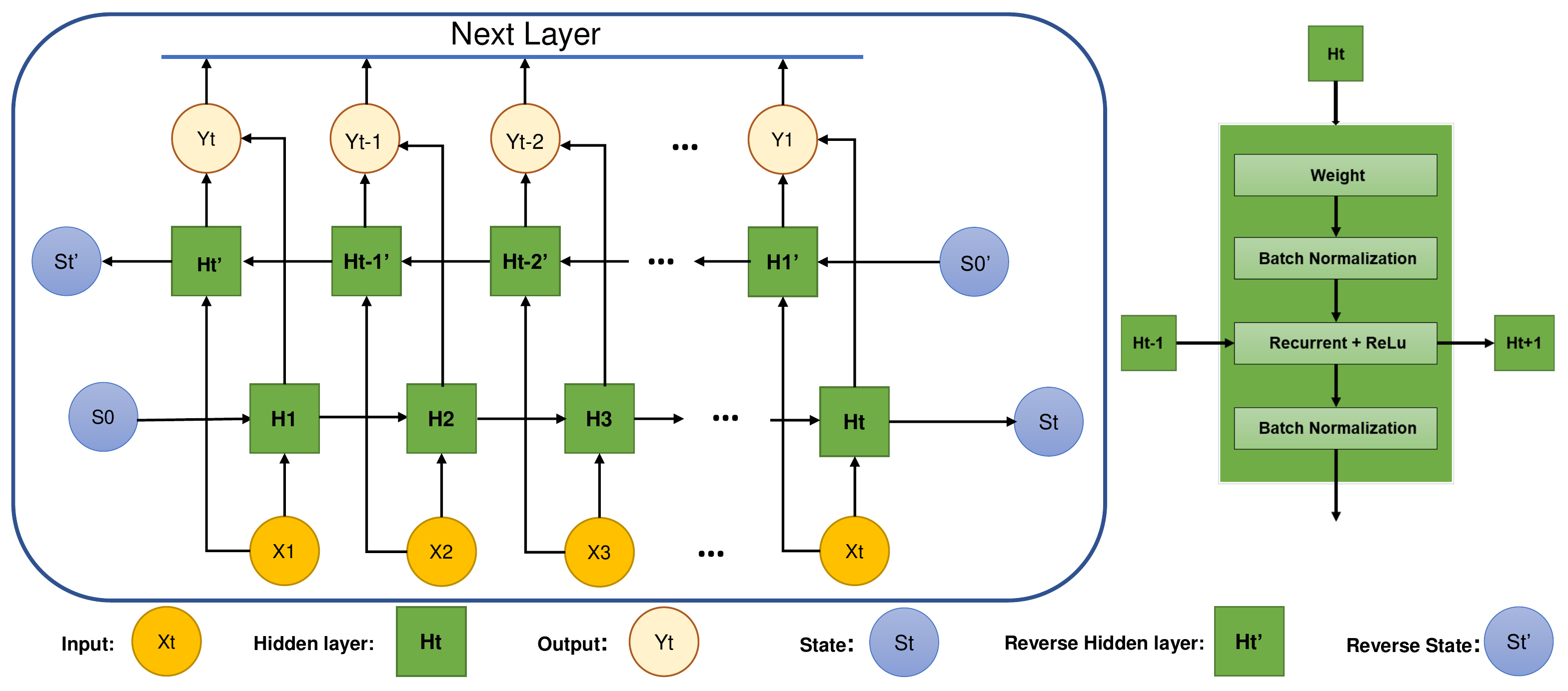}
	\caption{The proposed BML-indRNN model for surgical gesture recognition.}
	\label{fig:overviewINDrnn}
\end{figure*}

%Recurrent neural network (RNN) has been widely used for segmentation and classification of sequential data. In traditional RNN, the last hidden state is the input to the next state, which acts as a form of connection. We aim to build an RNN model to learn the underlying temporal features from the sequences of motion data for surgical gesture recognition. 

In most of the RNN based methods for sequential data processing, LSTM units  are widely used. For the applications of surgical workflow analysis, the length of the surgical operation videos normally has more than 1000 frames. Under this condition, traditional RNN methods with LSTM units may suffer from gradient decay over layers. Therefore, an improved RNN based model is required to develop for surgical gesture recognition.

%By modifying the hidden state connection mode from a fully connected mode, the transmission is conducted between single neurons. Therefore, 

The concept of an Independently recurrent neural network (IndRNN) has been proposed in \cite{li2018independently}. The hidden state $\mathbf{h_{t-1}}$ and $\mathbf{h_t}$ at timestep $t-1$ and $t$ are independent of each other, which provides better interpretability of the spatial features ($\mathbf{W}$) and temporal features ($\mathbf{u}$). Multiple layers can be stacked together to conduct correlation among different neurons. Therefore, each layer can process the outputs of all the neurons in the previous layer.  

We aim to apply the Independently RNN (IndRNN) cell \cite{li2018independently} to construct the model of BML-indRNN.  The IndRNN cell replaces the matrix multiplication (Eq.~\ref{RNN hidden state}) by element-wise vector multiplication  to calculate the hidden state, which means that each neuron has a single recurrent weight connected to its last hidden state.  
\begin{equation}
\mathbf{h_t}  = \sigma (\mathbf{W}\mathbf{x_t} + \mathbf{U}\mathbf{h_{t-1}} + \mathbf{b}) \label{RNN hidden state}
\end{equation}

The mathematical expression of the IndRNN is as follows:
\begin{equation}
	\label{indRNN hidden state}
\mathbf{h_t} = {\sigma}(\mathbf{W}\mathbf{x_t} + \mathbf{u} {\odot}\mathbf{h_{t-1}}  + \mathbf{b} )
\end{equation}
where $\odot$ represents Hadamard product, while the traditional operation is a dot product.  $\mathbf{x_t} \in \mathbb{R}^M$,  $\mathbf{h_t} \in \mathbb{R}^N$ are the input and hidden state at time step $t$.  $\mathbf{W} \in \mathbb{R}^{N{\times}M}$,  $\mathbf{u} \in \mathbb{R}^{N}$,  and $\mathbf{b} \in \mathbb{R}^N$ are the weights for the current input and the recurrent input, and the bias of the neurons. $\sigma(.)$ is an element-wise activation fuction of the neurons, and $\mathbf{N}$ is the number of neurons in the RNN layer.

To further improve the performance of the neural network for surgical gesture recognition, we extend the original independent RNN concept to a Bidirectional Multi-Layer indRNN (BML-indRNN) model.  The temporal features are considered when the sequence is reversed. The output at time $t$ depends not only on the information at the previous moment but also on the future moment.  The basic concept is to put two IndRNNs together by feeding the input time-sequential data in normal time order for one network and in reverse time order for another, which utilizes both  forward  and backward information about the sequence at every time step. In this way, the neural network can extract features in the forward and reverse order respectively. The architecture of the BML-indRNN model is shown in Fig. \ref{fig:overviewINDrnn}.  BML-indRNN model is constructed by three layers, including 64 hidden units in this paper.

\section{Experiments Results and Analysis}

\subsection{Evaluation Metrics}

Suppose that there are $n$ targeted classes, $C=[i,j]$ represents the confusion matrix with the dimension of $n{\times}n$. Each element of the confusion matrix represents that the sample from class $i$ is predicted as class $j$. Let $TP$ denotes True Positive, $TN$ denotes True Negative, $FP$ denotes False Positive, $FN$ denotes False Negative in the classification setting. Accuracy ($acc$), precision ($precision$), recall ($recall$) and F1-score ($F_{1}$) can be obtained as follows.

\begin{equation}
	\begin{cases}
		acc=(TP+TN)/(TP+FN+FP+TN) \\
		precision=TP/(TP+FP) \\
		recall=TP/(TP+FN)\\
		F_{1} =\frac{2}{\frac{1}{precision}+ \frac{1}{recall}} \\
	\end{cases}
\end{equation}

A confusion matrix can be generated for the evaluation of the multi-class classification setting, while Micro average and Macro average can be computed as follows for evaluations of the proposed method.

\begin{equation}
	\begin{cases}
	Micro = \frac{\sum_{i=1}^{n}C[i,i]}{\sum_{i,j=1}^{n}C[i,j]};\\	
Macro = \frac{1}{n}\sum_{i=1}^{n}\frac{C[i,i]}{\sum_{j=1}^{n}C[i,j]}
	\end{cases}
\end{equation}

The Micro average is computed as the average of total correct predictions across all classes. Macro average represents the average $TP$ rates for each class.  Macro F1-score ($F_{m1}$) is defined as the mean of class-wise F1-scores, while Macro Recall and Macro precision can be obtained for evaluation as well.

\subsection{Experimental Results}

%The proposed algorithm was implemented in Python using TensorFlow and Keras, and was trained on a PC with an Intel Core i5-8300H CPU (2.30 GHz), a GeForce GTX 1050 GPU (NVidia Corporation), and 8 GB of RAM.

The proposed algorithm was implemented in Python using TensorFlow and Keras, and was trained on a PC with a GeForce GTX 1050 GPU (NVidia Corporation) and 8 GB of RAM.
 %66,283 data points are used for training, while 28,373 data points are used for cross-validation during the training process. 

\subsubsection{Results for Spatial Feature Extraction Model Training}
The spatial feature extraction model is trained at first, with SGD as the optimizer and the loss function constructed by categorical cross-entropy. The batch size is set as 10. The learning rate decreases during the model training process, an exponential decay function is applied to reduce the learning rate as the training progresses. 

The data for experimental validation is divided using Leave One Trial Out (LOTO) for all the experiments, the setup of which is the same as described in \cite{ahmidi2017dataset}. The testing database includes 37086 data points.  As for the training database,   70\% of data is used for training, while the remaining 30\% of data is used for validation.   The training accuracy is 0.977, while the testing accuracy is 0.693 for training the spatial feature extraction model.

%After model training, unseen data divided via the Leave One Trial Out (LOTO) manner is used to test the performance of the model. The training accuracy is 0.977, while the testing accuracy is 0.693.

\begin{figure*}[tb]  
	\centering
	\includegraphics[width = 0.95\hsize]{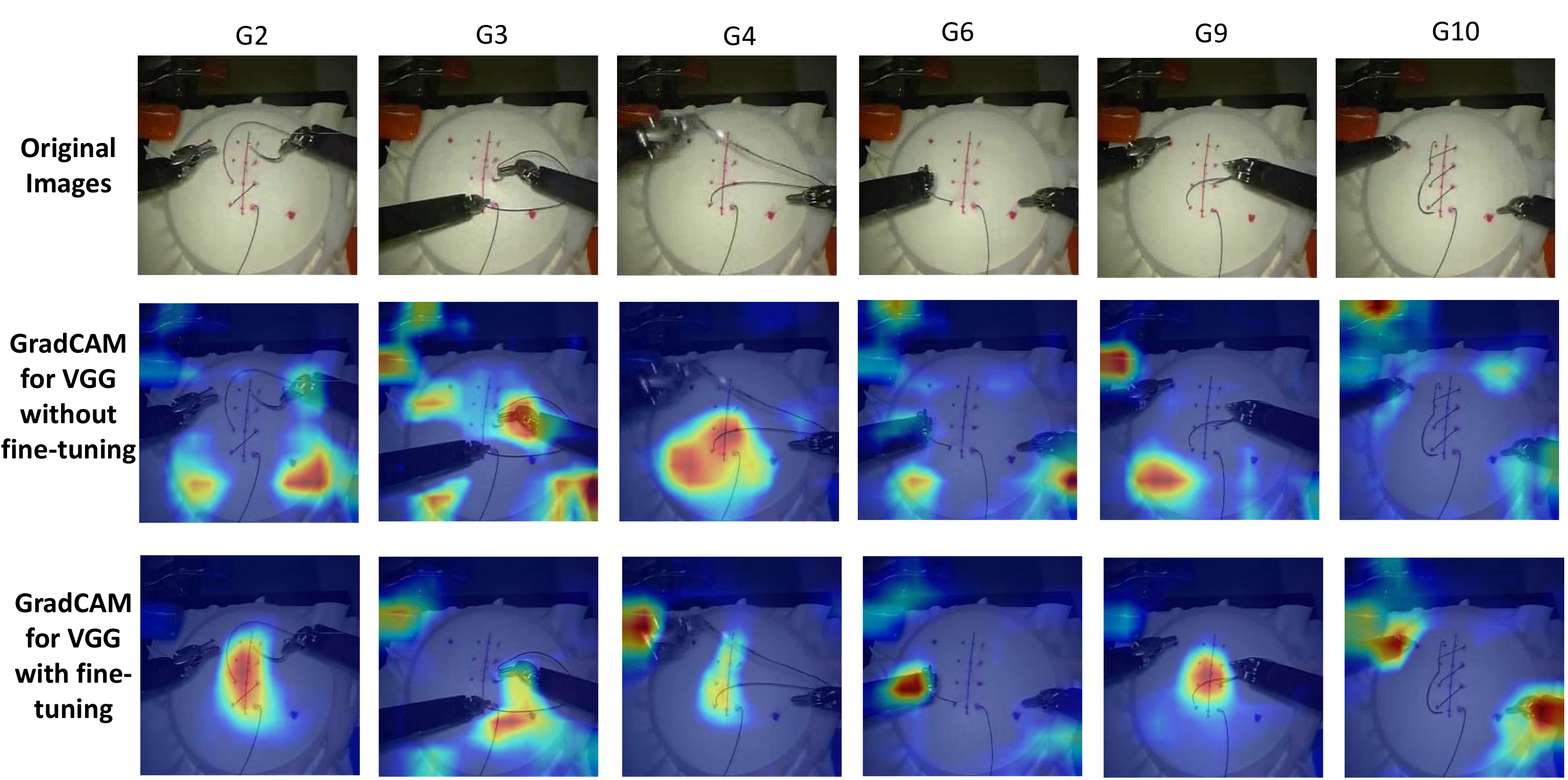}
	\caption{Six examples of the analysis of the spatial feature extraction model using GradCAM and the comparison between model with and without fine-tuning.}
	\label{fig:XAI.pdf}
\end{figure*}

\subsubsection{Results for Interpretability}
Fig. \ref{fig:XAI.pdf} demonstrates the explainable results for evaluating the Spatial Feature Extraction Model based on Grad-CAM. Image data selected from six different types of gesture classes is used as examples. For a particular category, the red color highlight the discriminative image regions used by the DCNN to identify that specific class of surgical gesture. 

Comparisons are made between the explainable visualization results of the model with and without fine-tuning. It can be clearly seen that without fine-tuning, the highlighted area for determining the gesture class is random, which means that the DCNN generates the results by reasoning some areas that do not have a close relationship with the operation scene. As for the explainable visualization results demonstrated in the third row, it can be clearly seen that the highlighted areas are related to the states of the suture, the needles and the tooltip of the surgical tools, which have a close relationship with the contexts of operation scenes and the corresponding gesture type.

\subsubsection{Results for BML-indRNN Model Training}

\begin{table}[!htb]
	\centering
	\caption{Surgical Gesture Recognition Results Based on the BML-IndRNN model}
	\label{tab:Results-RNN1}
	\begin{tabular}{c|c|c}
		\hline\hline
		& 	\textbf{Without Tuning}  	& \textbf{With Tuning}  \\\hline
		\textbf{Micro}  &  83.15\%  &   88.95\% \\
		\textbf{Macro} $\pm$ \textbf{std } &  73.41\%$\pm$19.70\% &    86.40\%$\pm$8.38\% \\
		\textbf{Precision} $\pm$ \textbf{std } &  69.68\%$\pm$2.49\% &  84.59\%$\pm$1.20\% \\
		\textbf{Macro Recall}  &  0.667 &  0.864 \\
		\textbf{Macro F1-Score}  &  0.666 &  0.849 \\
		\hline\hline
	\end{tabular}
\end{table}

\begin{figure}[tb]
	\centering
	\includegraphics[width = 0.9\hsize]{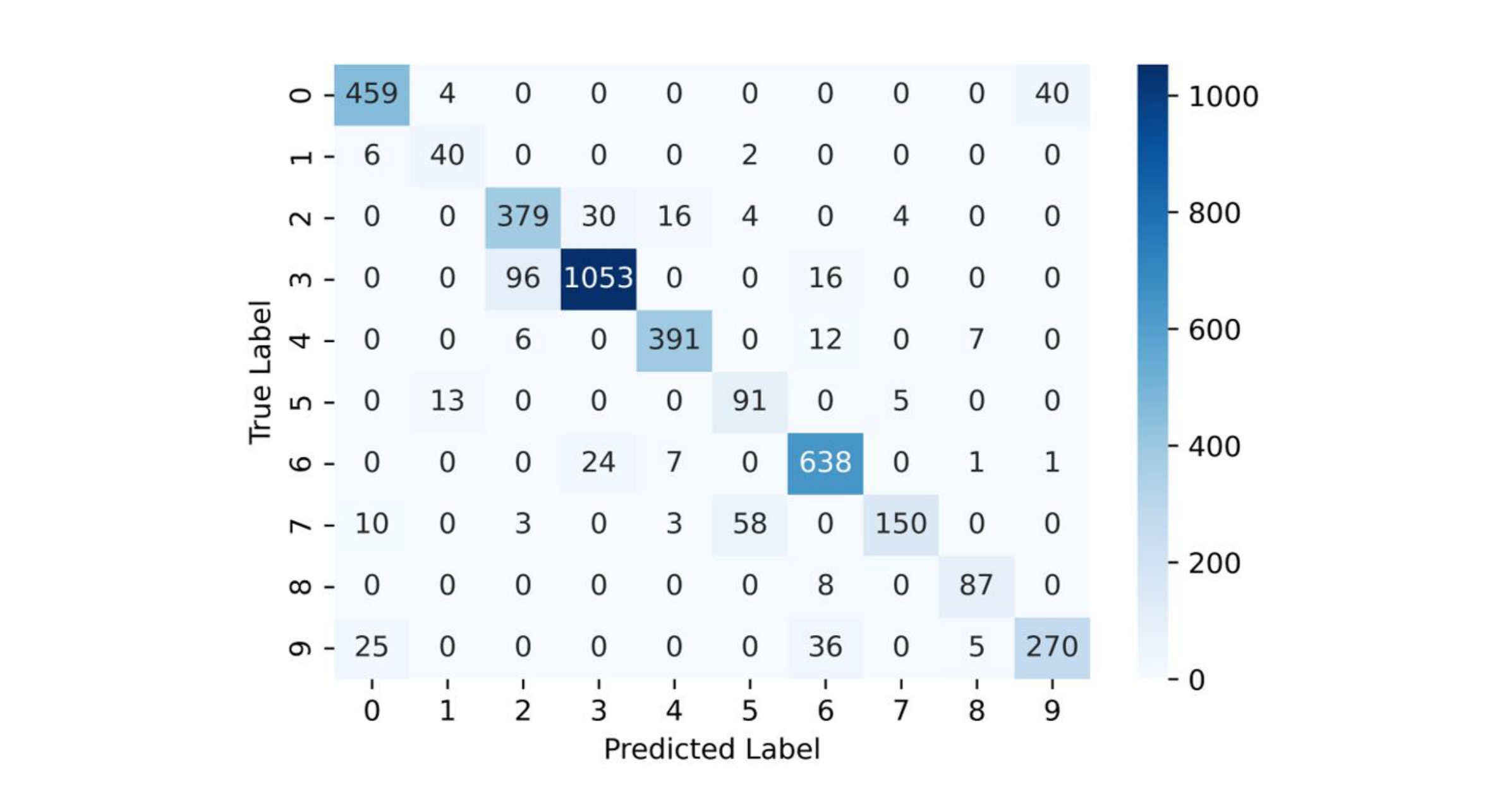}
	\caption{The confusion matrix for the surgical gesture classification results of the suturing task. }
	\label{fig:indRNN}
\end{figure}

The results for surgical gesture recognition using the proposed BML-indRNN are summarized in Table \ref{tab:Results-RNN1}, while the confusion matrix for the suturing task is calculated, as shown in Fig. \ref{fig:indRNN}.   The hyperparameters can be tuned for different surgical tasks to reach the desired performance.

\begin{figure}[tb]
	\centering
	\includegraphics[width = 0.7\hsize]{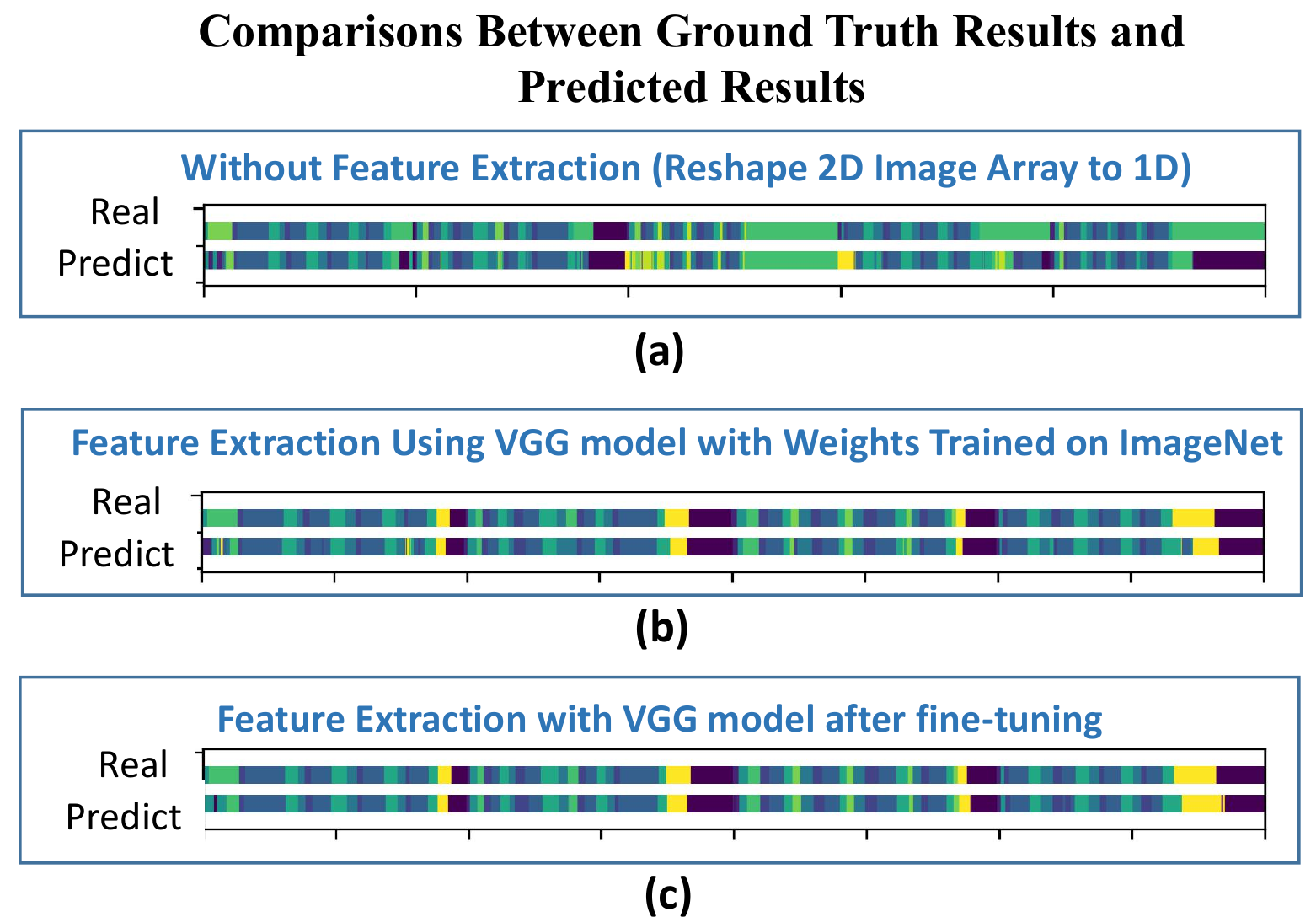}
	\caption{The visualization of comparisons between ground-truth data and predicted results. (a) Results predicted by model trained without DCNN based feature extraction. (b) Results predicted by model trained with VGG based model without fine-tuning. (c) Results predicted by model trained with VGG based model with fine-tuning.}
	\label{fig:ThreeColorBin.pdf}
\end{figure}

To verify the effectiveness of the spatial feature extraction model, comparisons are made between with and without using the spatial feature extraction model as well as with and without fine-tuning. The color-coded ribbon illustration for the comparisons between the ground-truth data and the prediction results are shown in Fig. \ref{fig:ThreeColorBin.pdf}.  Different colors represent different surgical gestures in the entire surgical procedure. The top line represents the ground-truth and the bottom line represents the predicted results.

The 2D image array obtained from video  can be reshaped to a 1D vector and then be fed into the BML-indRNN model for training. In this case, no feature extraction procedure is conducted for model training, while the prediction results are shown in Fig. \ref{fig:ThreeColorBin.pdf} (a), which can be regarded as the baseline for a comparative study. As indicated by the results in Fig \ref{fig:ThreeColorBin.pdf} (b) and (c), we can conclude that  fine-tuning can improve surgical gesture recognition performance.  

\subsubsection{Comparisons with the State-of-the-Art}

\begin{table}[htb]
	\centering
	\caption{The results for comparisons between the proposed methods and the baseline methods. }
	\begin{tabular}{cc|cc}
		\hline	\hline
		\textbf{Method }   & \textbf{Accuracy} &\textbf{Method }    & \textbf{Accuracy}   \\ \hline
		MsM-CRF             & 0.718  & Seg-ST-CNN     & 0.742\\ \hline
		TCN                 & 0.814  &	TCN+Deep RL         & 0.814 \\ \hline
		\textbf{BML-indRNN}& 0.822 & \textbf{BML-indRNN+DCNN}& 0.871 \\ \hline	\hline
	\end{tabular}
	\label{table:compare_suturing}
\end{table}

Markov/semi-Markov conditional random field (MsM-CRF)  model \cite{tao2013surgical}, Segmental
spatiotemporal convolutional neural network (Seg-ST-CNN) \cite{lea2016segmental}, Temporal convolutional networks (TCN) \cite{lea2016temporal}, and together with Deep Reinforcement Learning (DRL) \cite{liu2018deep} have been evaluated on the suturing task using JIGSAWS. All these methods can be used as the baselines for the performance evaluation of the surgical gesture recognition for the suturing task.

Table. \ref{table:compare_suturing} shows the results of the comparisons between the state-of-the-art algorithms and the proposed method in terms of the accuracy for surgical gesture recognition.  Results indicate that the proposed method \textbf{BML-indRNN+DCNN} outperforms the others.

% Comparisons are made between the state-of-the-art methods mentioned above and our proposed methods.

%Extended studies are carried out to verify the significance of model parameter fine-tuning.
%with an explainable spatial feature extraction
\section{Conclusions and Future Works}
In this paper,  a BML-indRNN  method is proposed for surgical gesture recognition. The proposed method is verified based on the video data of the suturing task obtained from JIGSAWS.  To demonstrate the significance of the spatial feature extraction model, comparisons are made between with and without using the DCNN based spatial feature extraction model to process the image data before feeding the data into the BML-indRNN model.  Results indicate that the classification accuracy can improve 4.97\% when using the DCNN model with fine-tuning for spatial feature extraction. The interpretability of the feature extraction model can be verified through GradCAM for post-hoc analysis. The proposed BML-indRNN is effective, since the overall accuracy of the surgical gesture recognition for the suturing task can reach 87.13\%, which outperforms several state-of-the-art approaches. 

Future work will include incorporating kinematic data to further enhance the classification accuracy for surgical gesture recognition.  The real-time performance of the model will be tested while more experimental validation can be conducted based on other surgical tasks, including needle passing tasks, knot-tying tasks, etc. By training effective models for surgical gesture segmentation and recognition, context-awareness for surgical operation assistance can be realized, while more reliable skill evaluation can be achieved by detailed analysis of the surgical gestures. 

%Bibliography
\bibliographystyle{unsrt}  
\bibliography{references}

\end{document}